\documentclass{article}
\usepackage[T1]{fontenc}
\usepackage{spconf,amsmath,graphicx}

\usepackage{enumitem}
\setlist{nosep, leftmargin=14pt}

\usepackage{mwe} 

\usepackage{booktabs}
\usepackage{adjustbox}
\usepackage{xcolor}
\usepackage{colortbl}
\usepackage{arydshln}
\usepackage{makecell}
\usepackage{fontawesome5}
\usepackage[numbers]{natbib}
\renewcommand{\paragraph}[1]{\vspace{1em}\noindent\textbf{#1} }
\definecolor{gold}{RGB}{212,175,55}
\definecolor{silver}{RGB}{192,192,192}
\definecolor{bronze}{RGB}{205,127,50}
\title{Benchmarking VLM Medical Hallucination across 5 Modalities}

\usepackage{url}
\usepackage{cleveref}
\usepackage{pifont}
\newcommand{\cmark}{\ding{51}}
\newcommand{\xmark}{\ding{55}}

\name{Author(s) Name(s)\thanks{Some author footnote.}}
\address{Author Affiliation(s)}
\twoauthors
 {Zhixing Cao\textsuperscript{*}}
	{Technical University of Munich\\
	TUM School of Natural Sciences}
 {Yusen Peng\textsuperscript{*}}
	{The Ohio State University\\Computer Science and Engineering}

\begin{document}
%
\maketitle
\begin{abstract}
Hallucinations in Vision-Language Models (VLMs) within the medical domain remain a difficult and not well-understood problem, and it is still unclear where these models break down. Although several benchmarks targeting medical hallucination have been introduced, they often lack broad coverage across different medical imaging modalities. We introduce Multi-Med, a medical hallucination benchmark containing 1,000 image–question pairs spanning five distinct modalities. Using this benchmark, we perform an in-depth experimental study of 10 widely used VLMs across five modalities and four question categories. Our findings indicate that model performance differs substantially depending on the imaging modality and question type, and that the differences among the best-performing models are not statistically significant.
\end{abstract}

\section{Introduction}
\label{sec:intro}
Vision-Language Models (VLMs)~\citep{llava, qwenvl}, which produce text conditioned on both images and textual prompts, have achieved strong results across many visual question answering benchmarks, including those requiring reasoning~\citep{MMMU} and OCR-centric understanding~\citep{TextVQA, OCRBench}. To better match these capabilities to medical settings, several medical-domain finetuned models have also been proposed~\cite{li2023llavamedtraininglargelanguageandvision, sellergren2026medgemmatechnicalreport}. However, hallucinations in the medical domain remain a persistent challenge, and it is still not well understood where these models tend to fail. As a result, multiple benchmarks centered on medical hallucination have been developed~\cite{gu2024medvhsystematicevaluationhallucination,chang2025medhevalbenchmarkinghallucinationsmitigation,chen2026detectingevaluatingmedicalhallucinations}, along with detailed analyses and broad model coverage.

However, current benchmarks for medical hallucinations still lack sufficient \emph{modality coverage}: medical images may come from many different sources, including CT, MRI, X-ray, ECG, EEG, and even pathology imaging. MedVH~\cite{gu2024medvhsystematicevaluationhallucination} evaluates models only on chest X-ray (CXR), whereas Med-HallMark~\cite{chen2026detectingevaluatingmedicalhallucinations} and MedHEval~\cite{chang2025medhevalbenchmarkinghallucinationsmitigation} extend coverage by adding two more modalities, CT and MRI. Therefore, to achieve broader coverage, we introduce a benchmark spanning five modalities by additionally incorporating \textbf{ECG} and \textbf{Pathology} images during dataset construction. We also emphasize that preventing an imbalanced modality mix in the question set is equally crucial. Thus, throughout curation, we keep the number of questions per modality (CT, MRI, CXR, ECG, pathology) approximately equal to maintain a balanced distribution. In summary, our contributions are threefold:

\begin{enumerate}
    \item We assemble a medical hallucination benchmark with 1,000 image–question pairs spanning five modalities, including two newly added ones: ECG and pathology;
    \item Using our benchmark, we perform an in-depth experimental evaluation of 10 widely used vision-language models, covering five modalities and four question types to create a testbed rich in modalities and diverse in scenarios;
    \item We find that performance gaps among top models are not statistically significant; Hallucination accuracy generally scales with model size, though parameter efficiency varies; Performance varies widely by image modality and question type; At the same scale, domain-specialized models lag behind general-purpose ones.
\end{enumerate}

\section{Related Work}
\begin{table}[ht]
    \centering
    \vspace{-10pt}
    \adjustbox{width=0.48\textwidth}{
    \begin{tabular}{cccccc}
        \toprule
        \textbf{\thead{Benchmarks}} & \textbf{\thead{Modality\\Coverage}} & \textbf{\thead{\#models}} & \textbf{\thead{balanced\\control}} & \textbf{\thead{curation\\process}} \\
        \midrule        MedVH(2024)~\cite{gu2024medvhsystematicevaluationhallucination} & \cellcolor{red!10} CXR & 8 & \xmark & \xmark \\ 
        Med-HallMark(2024)~\cite{chen2026detectingevaluatingmedicalhallucinations} & \cellcolor{yellow!10}CT,MRI,CXR & 10 & \xmark & \cmark  \\
        MedHEval(2025)~\cite{chang2025medhevalbenchmarkinghallucinationsmitigation} & \cellcolor{yellow!10}CT,MRI,CXR & 11 & \xmark & \cmark \\
        \hdashline
        \textbf{MultiMed(Ours)} & \cellcolor{green!10}\makecell{CT,MRI,CXR,\\ECG,Pathology}  & 10 & \cmark & \cmark \\ 
        \bottomrule
    \end{tabular}}
    \vspace{-10pt}
    \caption{Benchmarks comparison.}
    \vspace{-20pt}
    \label{tab:comparison}
\end{table}
\paragraph{Medical hallucination benchmarks.}
Several benchmarks have been developed to evaluate the medical hallucination behavior of VLMs, as illustrated in \cref{tab:comparison}. MedVH~\cite{gu2024medvhsystematicevaluationhallucination} examines hallucinations through various tasks involving mismatched images, invalid answer choices, and incorrect clinical premises, but its primary diagnostic evaluation is limited to chest X-rays. Med-HallMark~\cite{chen2026detectingevaluatingmedicalhallucinations} and MedEval~\cite{chang2025medhevalbenchmarkinghallucinationsmitigation}, extend the evaluation to multiple modalities including CXR, CT, and MRI. Unlike MedVH, they also make effort in curating questions instead of blindly sampling from existing dataset archives. Despite all these efforts, the existing benchmarks are still dominated by conventional \emph{radiological images} and subject to \emph{imbalanced distributions} across all modalities. To address these two limitations, we present MultiMed, a VLM medical hallucination benchmark across five modalities, each of which is provided with a balanced number of questions.

\section{Multi-Med Benchmark}
\begin{figure}[ht]
    \centering
    \vspace{-10pt}
    \includegraphics[width=1.0\linewidth]{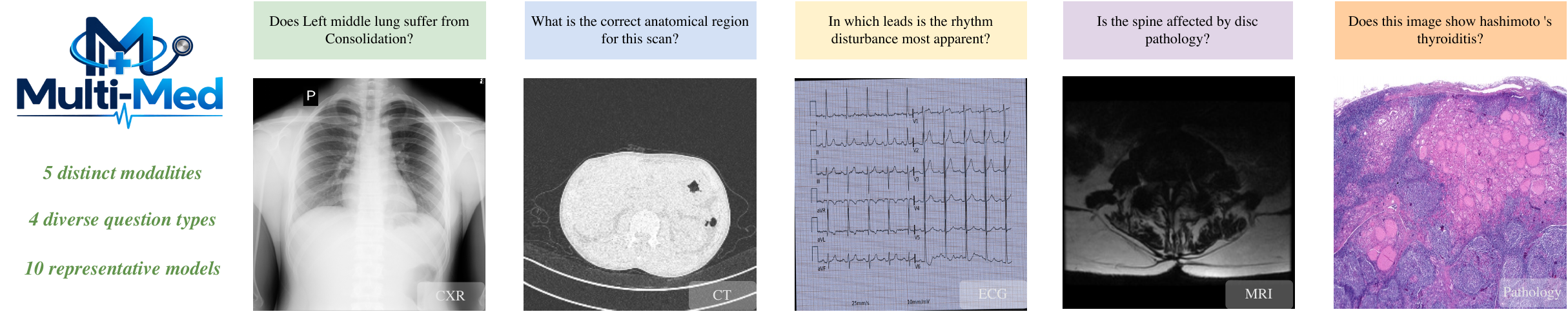}
    \vspace{-20pt}
    \caption{Benchmark overview of Multi-Med.}
     \vspace{-20pt}
    \label{fig:benchmark}
\end{figure}

\paragraph{Benchmark Overview.} As illustrated in \cref{fig:benchmark}, Multi-Med provides a unified framework for evaluating medical VLMs across five different modalities beyond radiological images like CXR, CT, and MRI, including physiological signal plots like ECG and pathology images like microscopic tissue images and other clinical images. MultiMed evaluates ten representative VLMs using four different question types: baseline, mismatch, incorrect premise, and false suggestion. This design enables us to examine whether model performance and hallucination robustness generalize across distinct \emph{visual representations} and how models respond to different \emph{forms of misleading input}. Multi-Med enables a controlled analysis of modality-specific hallucinations and model sensitivity to different forms of hallucination scenarios.

\paragraph{Benchmark Composition.} Multi-Med consists of 1,000 image-related question pairs covering five medical modalities, as shown in \cref{tab:details}: CT, MRI, CXR, ECG, and pathology. The source images were collected from existing public archives and benchmark datasets~\citep{butsanets2026radimagenetvqalargescalectmri, Nguyen_2025, liu2024teachmultimodalllmscomprehend, he2020pathvqa30000questionsmedical}.
The image-question pairs were then curated into four categories: baseline questions for assessing standard medical image understanding; modality mismatch questions for evaluating image–question alignment; incorrect-premise questions for testing the recognition of false assumptions; and false-suggestion questions for measuring resistance to misleading textual guidance. We illustrate our question curation pipeline in the next \cref{sec:curation}. Each question type contains exactly 250 samples, while each modality contributes between 192 and 222 samples. This \textbf{balanced} distribution enables \emph{controlled studies and fair comparisons} among different hallucination conditions while supporting modality-specific analysis.
\begin{table}[ht]
    \centering
    \vspace{-10pt}
    \adjustbox{width=0.48\textwidth}{
    \begin{tabular}{ccccc|c}
        \toprule
        \textbf{\thead{Modalities}} & \textbf{\thead{Baseline}} & \textbf{\thead{Modality\\Mismatch}} & \textbf{\thead{Incorrect\\Premise}} &  \textbf{\thead{False\\Suggestions}} & \textbf{\thead{Total}} \\
        \midrule
        CT & 45 & 63 & 54 & 60 & 222 \\
        MRI & 45 & 44 & 52 & 54 & 195 \\
        CXR & 48 & 44 & 52 & 48 & 192 \\
        ECG & 67 & 56 & 43 & 31 & 197 \\
        Pathology & 45 & 43 & 49 & 57 & 194 \\
        \midrule
        Total & 250 & 250 & 250 & 250 & 1000 \\
        \bottomrule
    \end{tabular}}
    \vspace{-10pt}
    \caption{Question Details of Multi-Med.}
    \vspace{-20pt}
    \label{tab:details}
\end{table}

\begin{figure}[ht]
    \centering
    \vspace{-10pt}
    \includegraphics[width=1.0\linewidth]{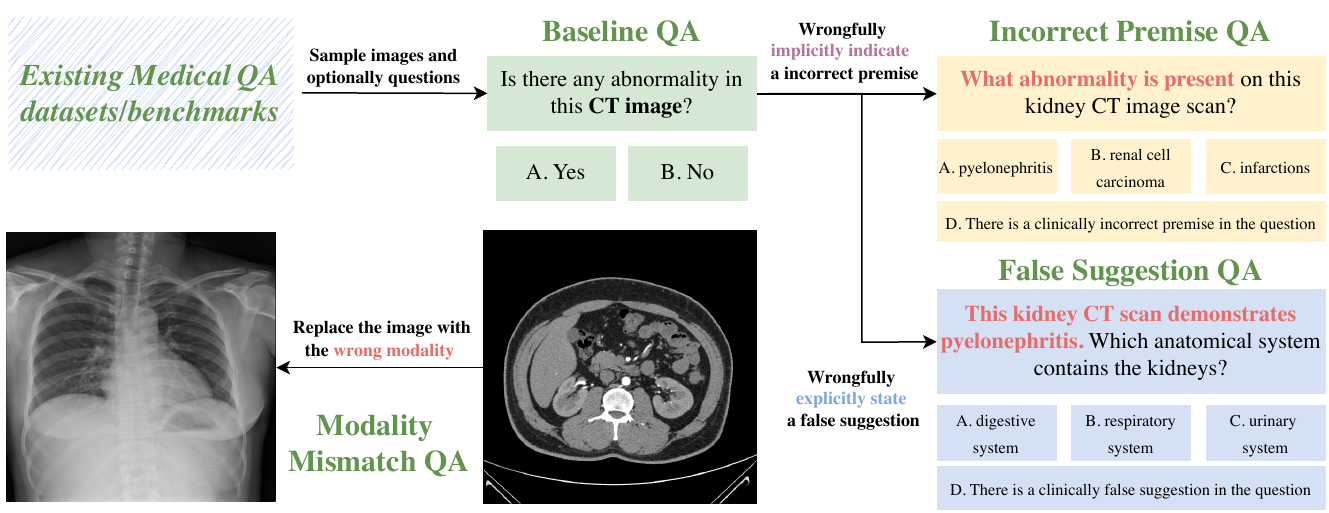}
    \vspace{-20pt}
    \caption{Question Curation Pipeline.}
    \vspace{-20pt}
    \label{fig:curation}
\end{figure}

\paragraph{Question Curation Pipeline.}\label{sec:curation} \cref{fig:curation} illustrates the question curation pipeline and explains how each question type is constructed and what capability it tests. We start by sampling medical images and, when available, compact their associated questions from existing datasets and benchmarks to create baseline questions that assess standard medical image understanding. Then, using these baseline questions as anchors, we construct from three hallucination-oriented question types. Modality-mismatch questions would pair a question with an image that is from an unrelated modality to evaluate image-question alignment. Incorrect-premise questions would introduce an unsupported assumption implicitly, which would examine whether models would verify the premise before answering the prompt. In contrast, false-suggestion questions would present the incorrect clinical claim explicitly to assess whether the models could resist misleading textual claims. All questions and answer choices are verified individually to ensure validity.
\section{Experiments}
In our evaluation study, we evaluate a total of ten representative VLMs from three different kinds: 1 Frontier close-source model (GPT-4o), 6 general-purpose open-source models (LLaVA 1.5 7B and 13B, Qwen2VL 2B and 7B, Qwen3VL 4B and 8B), and 3 medical-domain models (LLaVA-Med 7B, MedGemma 4B and 27B). We present aggregated accuracy by both question types and image modalities, as shown in \cref{tab:main}. We also report the overall aggregated score: Qwen3VL 8B reaches the highest overall accuracy 72.9\%, followed by GPT-4o with 70.2\% and Qwen3VL 4B with 64.2\%. In the following sections, we present analysis regarding the global performance landscape, tradeoff investigation between model size and accuracy, taxonomy-specific analysis across different different modalities and question types. We finally shed light on the comparison between general-purpose models and medical-specialized domain models.
\begin{table*}[ht]
    \centering
    \footnotesize
    \adjustbox{width=1.0\textwidth}{
    \begin{tabular}{ccccccccccc}
        \toprule
        \textbf{\thead{Models}} & \textbf{\thead{BL}} & \textbf{\thead{MM}} & \textbf{\thead{IP}} & \textbf{\thead{FS}} & \textbf{\thead{CT}} & \textbf{\thead{MRI}} & \textbf{\thead{CXR}} & \textbf{\thead{ECG}} & \textbf{\thead{Pathology}} & \cellcolor{yellow!30}\textbf{\thead{Overall}} \\
        & \multicolumn{4}{c}{\thead{by question types}} & \multicolumn{5}{c}{\thead{by modalities}} & \\
        
        \midrule
        \rowcolor{orange!10}\multicolumn{2}{c}{\textit{\textbf{Frontier close-source models}}} & & & & & & & & & \\
        \textcolor{silver}{\faMedal} GPT-4o & 64.0 & 99.2 & 83.2 & 34.4 & 70.72 & 74.87 & 73.43 & 62.43 & 69.58 & \cellcolor{yellow!30}\textbf{70.2} \\        \rowcolor{green!10}\multicolumn{2}{c}{\textit{\textbf{General open-source models}}} & & & & & & & & & \\
        LLaVA-1.5-7B~\cite{liu2023visualinstructiontuning} & 42.8 & 48.8 & 52.0 & 17.6 & 42.34 & 32.30 & 43.75 & 42.13 & 40.72 & \cellcolor{yellow!30}40.3 \\
        LLaVA-1.5-13B~\cite{liu2023visualinstructiontuning} & 46.8 & 56.4 & 54.0 & 18.4 & 45.94 & 46.66 & 45.31 & 41.62 & 39.69 & \cellcolor{yellow!30}43.9 \\
        Qwen2-VL-2B~\cite{wang2024qwen2vlenhancingvisionlanguagemodels} & 51.2 & 24.0 & 10.8 & 0.8 & 19.82 & 31.79 & 14.06 & 27.91 & 14.94 & \cellcolor{yellow!30}21.7 \\
        Qwen2-VL-7B~\cite{wang2024qwen2vlenhancingvisionlanguagemodels} & 55.6 & 80.8 & 20.0 & 3.6 & 44.14 & 46.66 & 31.77 & 39.59 & 37.11 & \cellcolor{yellow!30}40.0 \\
        \textcolor{bronze}{\faMedal} Qwen3-VL-4B~\cite{bai2025qwen3vltechnicalreport} & 64.8 & 97.6 & 74.8 & 19.6 & 64.86 & 69.74 & 68.75 & 60.40 & 57.22 & \cellcolor{yellow!30}64.2 \\
        \textcolor{gold}{\faMedal} Qwen3-VL-8B~\cite{bai2025qwen3vltechnicalreport} & 62.8 & 98.8 & 92.0 & 38.0 & 72.07 & 82.05  & 77.08 & 64.46 & 69.07 & \cellcolor{yellow!30}\textbf{72.9} \\        \rowcolor{cyan!10}\multicolumn{2}{c}{\textit{\textbf{Domain open-source models}}} & & & & & & & & & \\
        LLaVA-Med-7B~\cite{li2023llavamedtraininglargelanguageandvision} & 22.8 & 4.4 & 0.0 & 2.8 & 8.10 & 12.30 & 4.16 & 6.59 & 6.18 & \cellcolor{yellow!30}7.5 \\
        MedGemma-4B~\cite{sellergren2026medgemmatechnicalreport} & 51.2 & 52.4 & 2.4 & 2.0 & 25.67 & 24.10 & 32.81 & 30.96 & 21.64 & \cellcolor{yellow!30}27.0 \\
        MedGemma-27B~\cite{sellergren2026medgemmatechnicalreport} & 51.2 & 86.0 & 70.0  & 21.2 & 63.96 & 62.05 & 60.93 & 49.74 & 47.93 & \cellcolor{yellow!30}\textbf{57.1}  \\
        \bottomrule
    \end{tabular}}
    \vspace{-10pt}
    \caption{\textbf{Evaluation results}. BL=baseline, MM=Modality mismatch, IP=Incorrect premise, FS=False suggestion.}
    \vspace{-15pt}
    \label{tab:main}
\end{table*}
\subsection{Global Performance Landscape}
\begin{figure}[ht]
    \centering
    \includegraphics[width=1.0\linewidth]{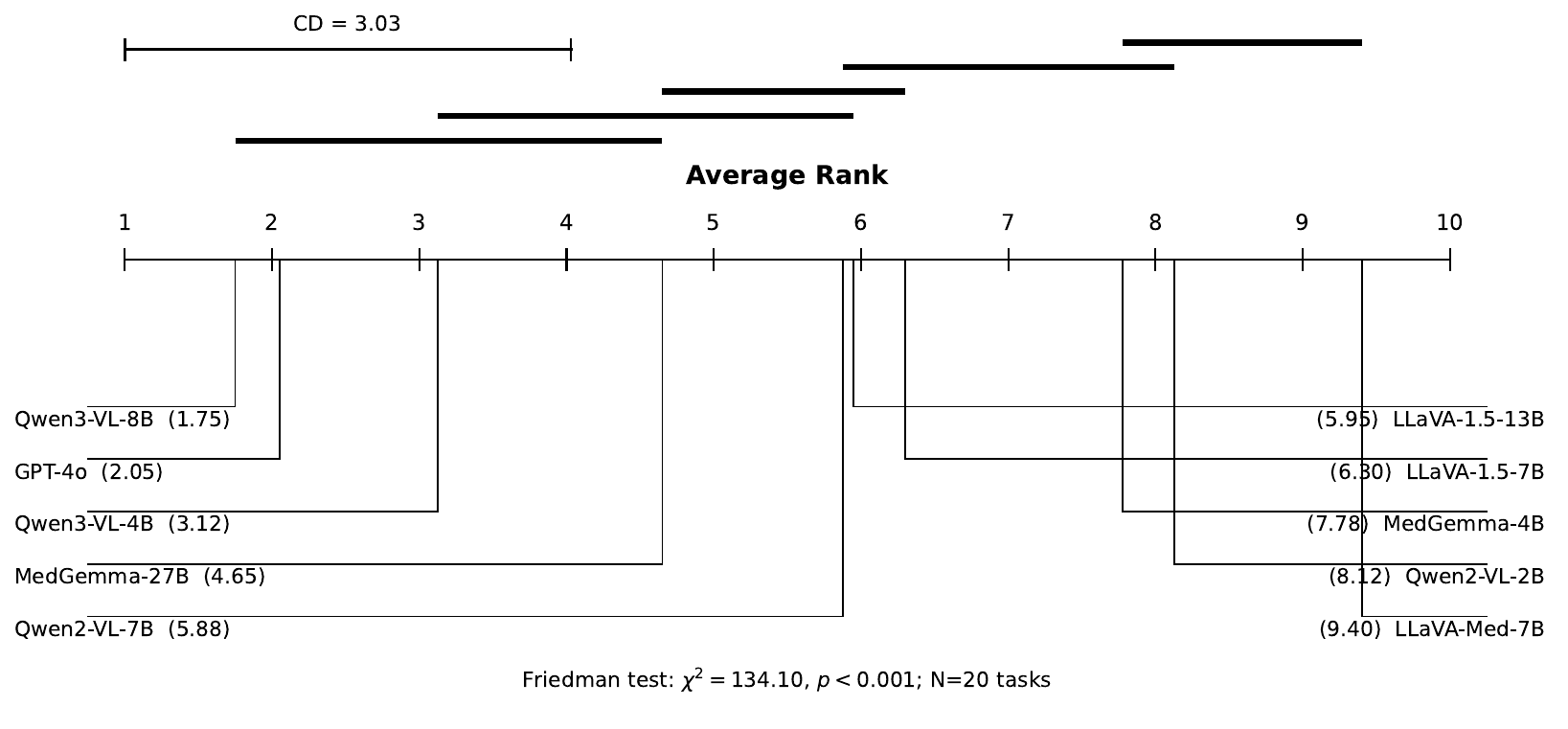}
    \vspace{-20pt}
    \caption{Average ranking CD diagram.}
    \vspace{-10pt}
    \label{fig:CD_diagram}
\end{figure}
\noindent To avoid drawing any conclusion from a single, heavily-aggregated accuracy score in \cref{tab:main}, we need a more reliable measure of the global performance landscape. Owing to the rich taxonomy of our proposed benchmark, we can define a total of 20 tasks given 4 different question types and 5 different image modalities. For each task, we rank all 10 models in our study and calculate the average ranking, illustrated in \cref{fig:CD_diagram}. We also apply Friedman-Nemenyi test to check statistical significance, in which the performance of models under the same bold line will not be statistically significant. From \cref{fig:CD_diagram}, we find that although Qwen3VL 8B and GPT-4o are dominating the leaderboard in \cref{tab:main}, their holistic performance is not statistically significant from Qwen3VL 4B and MedGemma 27B. This observation highlights the necessity of ranking models from a holistic and multi-faceted perspective.
\subsection{Model size-accuracy tradeoff}
\begin{figure}[ht]
    \centering
    \includegraphics[width=0.9\linewidth]{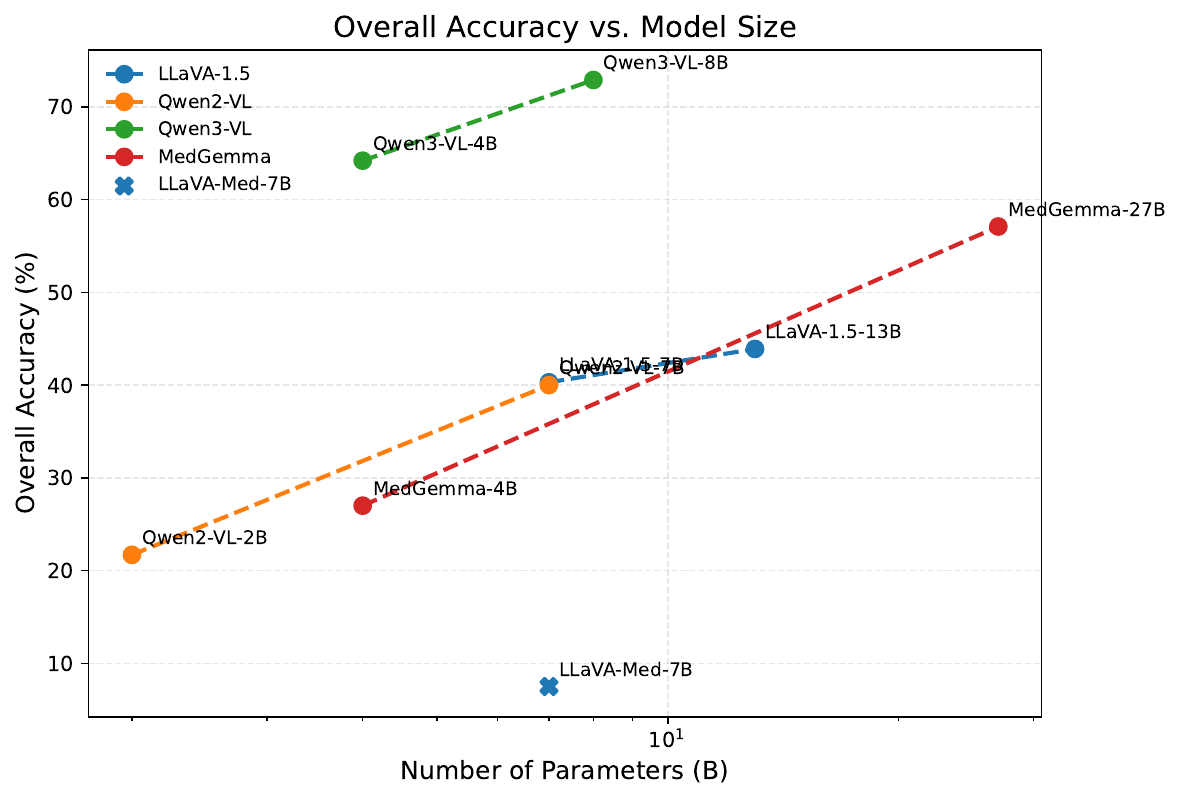}
    \vspace{-15pt}
    \caption{Model size-accuracy tradeoff.}
    \vspace{-15pt}
    \label{fig:tradeoff}
\end{figure}
\noindent It's equally important for an model to be parameter and memory efficient in addition to performing well on detecting hallucination. Thereby, we further investigate the tradeoff between hallucination detection accuracy and the model sizes of VLMs. In \cref{fig:tradeoff}, we present a scatter plot that demonstrates the tradeoff between overall accuracy and number of model parameters in a billion scale. We do not include GPT-4o since it's a close source model without information about model parameters. In \cref{fig:tradeoff}, we also connect models with different sizes but falling under the same model family to better illustrate their scaling behaviors.

\noindent We observe a very consistent scaling trend of increasing accuracy with more model parameters for all four model families: Qwen3VL, Qwen2VL, LLaVA, and MedGemma. However, the magnitude of performance gains vary a lot among different model families: scaling LLaVA from 7B to 13B only leads to marginal improvement, whereas scaling Qwen3VL from 4B to 8B brings greater performance boost. This implies that model architectures and training corpus may greatly affect how model scales. Nevertheless, different model families also exhibit very distinct parameter efficiency. For example, although MedGemma 27B is seven times larger than Qwen3VL 4B, its performance is still trailing Qwen3VL 4B. This is also in indicator that blindly scaling the model size can bring suboptimal performance outcomes, and more improvement on both model design and training strategy is needed.
\subsection{Taxonomy-specific analysis}
\begin{figure}[htb]
\begin{minipage}[b]{.48\linewidth}
  \centering
  \centerline{\includegraphics[width=4.0cm]{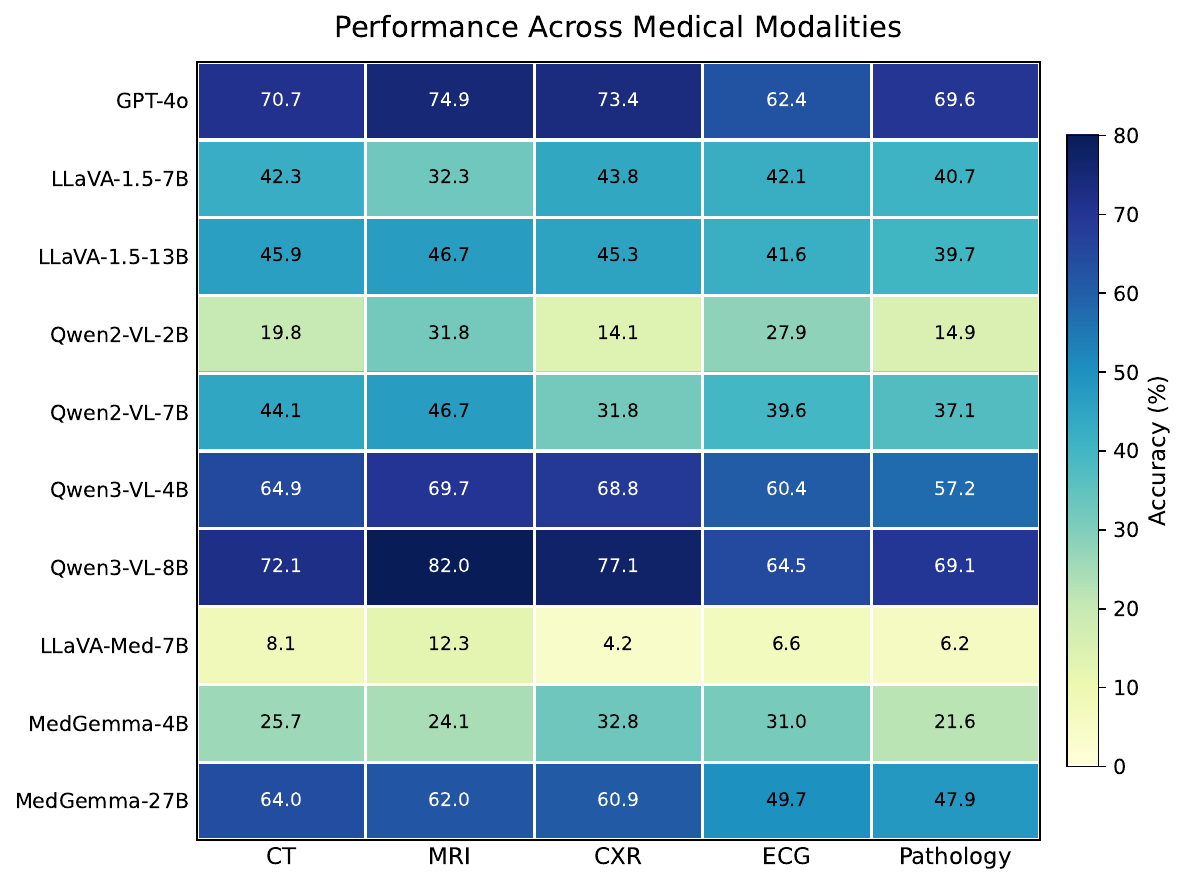}}
  \centerline{(a) modality heatmap}\medskip
\end{minipage}
\hfill
\begin{minipage}[b]{0.48\linewidth}
  \centering
  \centerline{\includegraphics[width=4.0cm]{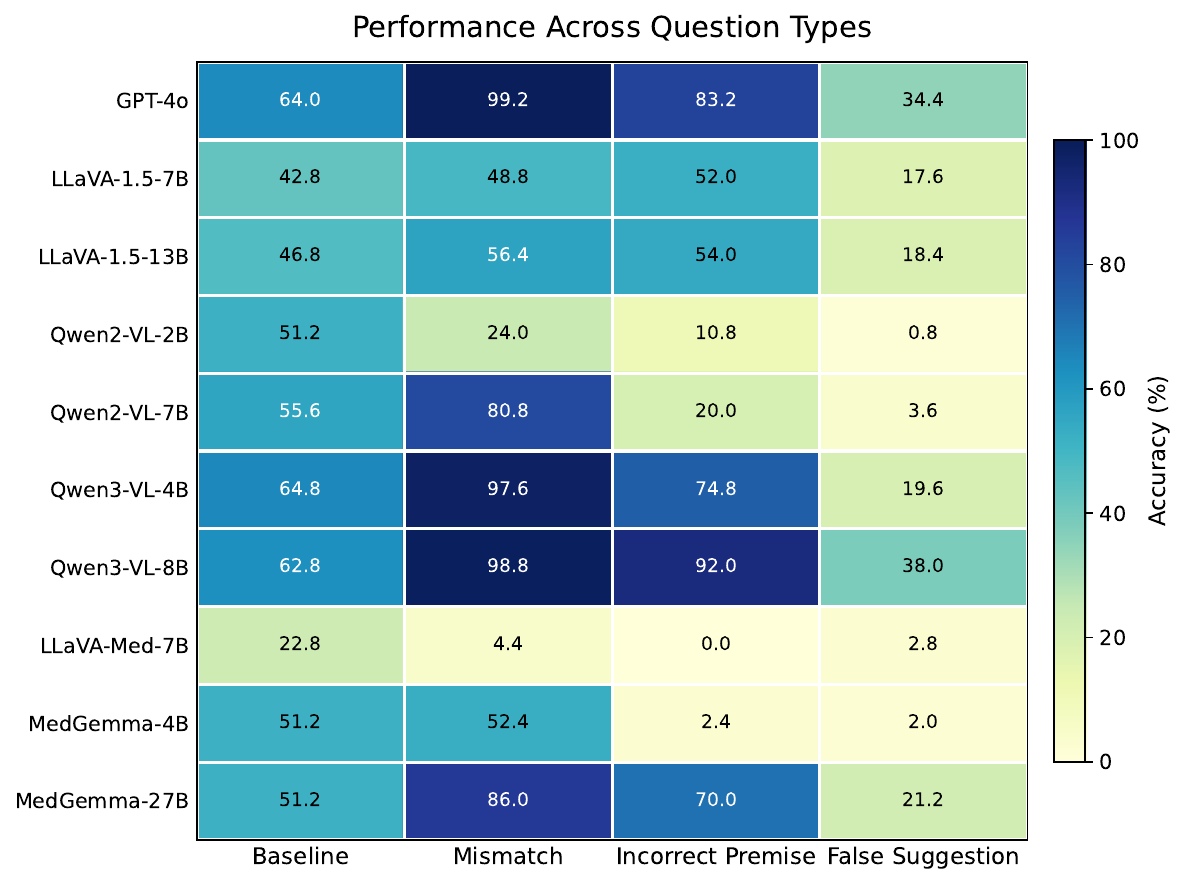}}
  \centerline{(b) question heatmap}\medskip
\end{minipage}
\vspace{-15pt}
\caption{Taxonomy-specific analysis.}
\label{fig:heatmaps}
\vspace{-20pt}
\end{figure}

\paragraph{Modalities.} 
To evaluate the models' performance in individual modalities, \cref{fig:heatmaps} (a) presents a heatmap of modality-specific accuracy, where darker blue indicates higher performance. Qwen3-VL-8B achieves the strongest overall results, leading on CT (72.1\%), MRI (82.0\%), CXR (77.1\%), and ECG (64.5\%), while GPT-4o performs consistently well and achieves the highest pathology accuracy (69.6\%). Across all models, the mean accuracies for CT, MRI, and CXR are 45.8\%, 48.3\%, and 45.2\%, respectively, compared with 42.6\% for ECG and 40.4\% for pathology: this pattern suggests that current VLMs generalize less effectively to \emph{physiological signals} such as ECG and microscopic pathology images than to conventional radiological modalities commonly represented in existing benchmarks. These results further highlight the importance of our work in which it includes ECG and pathology modalities in hallucination evaluation.

\paragraph{Questions.} 
\cref{fig:heatmaps} (b) compares model accuracy across the four question types. Similarly, darker blue indicates higher performance. It can be observed that each model performs really well in the Modality Mismatch question type: mismatch questions produce the highest mean accuracy (64.8\%). This task primarily requires models to recognize an incompatibility between the image and question rather than solve the underlying clinical problem. The baseline is slightly harder compared with the mismatch question, as it requires the model to actually use medical-related knowledge in order to get the question correct. It yields a mean accuracy of 51.3\% where Qwen3-VL-4B performs best at 64.8\%. The next two types of questions are slightly more challenging compared to the previous two, as expected. Incorrect-premise questions produce a lower mean accuracy of 45.9\% and show substantial variation across models, with Qwen3-VL-8B achieving the highest accuracy of 92.0\%. This indicates that only the strongest models consistently identify unsupported assumptions embedded implicitly in a question. False-suggestion questions are the most challenging overall, producing a mean accuracy of only 15.8\%, while the best-performing model, Qwen3-VL-8B, reaches just 38.0\%. These results suggest that current VLMs are particularly susceptible to explicit but unsupported clinical claims, which may override evidence from the accompanying image. 

\subsection{General v.s. Domain specialized models}
\begin{figure}[ht]
    \centering
    \includegraphics[width=1.0\linewidth]{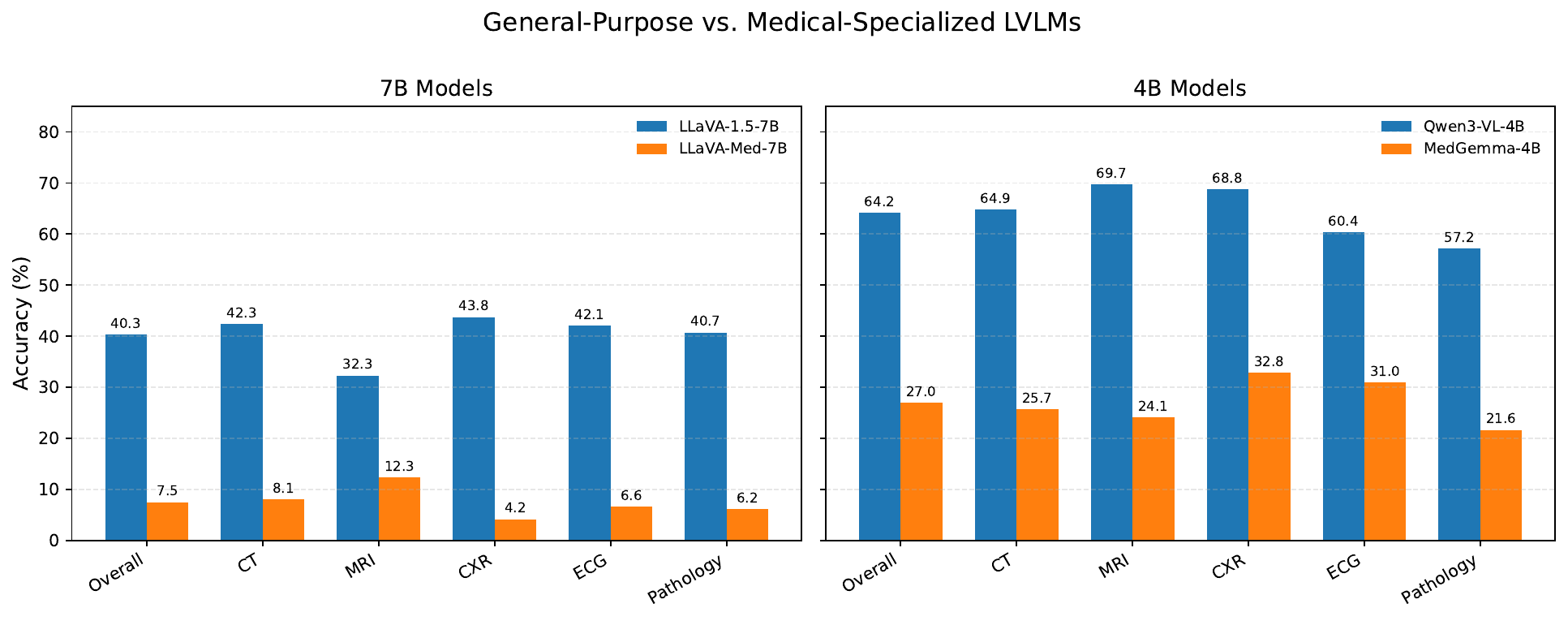}
    \vspace{-25pt}
    \caption{General vs. Domain models.}
    \vspace{-20pt}
    \label{fig:general_vs_domain}
\end{figure}

\noindent We further analyze the performance between general-purpose VLMs and medical-domain VLMs and compare models of similar size. In \cref{fig:general_vs_domain}, we compare two sets of such: for 7B models, we compare LLaVA-1.5 and LLaVA-Med; for 4B models, we compare Qwen3-VL and MedGemma. We report the overall accuracy on our benchmark and modality-specific performance as well. We observe that general-purpose models actually perform better under the same model size (7B and 4B) under every single modality, compared to medical-domain models. This sheds light on ``an illusion of progress'' of finetuning models on medical-domain data: supervised finetuning may help the model retain more medical knowledge, but it's equally important to preserve the model's robustness against hallucination scenarios.
\section{Conclusion}
In this study, we introduce a medical hallucination benchmark spanning five image modalities, expanded by incorporating images derived from physiological signals. Benchmarking ten widely used VLMs reveals practical observations: performance differs substantially across image modalities and question categories. We anticipate this benchmark will support future evaluations on an even broader range of modalities, enabling a more comprehensive understanding of hallucination behavior in modern vision-language models within medical settings.

\subsection*{Acknowledgment}
We use generative AI in 1) rewriting and curating some benchmark questions 2) polishing the manuscript. No funding was received for conducting this study. The authors have no relevant financial or non-financial interests to disclose.

\subsection*{Compliance with Ethical Standards}
This is a model benchmark study for which no ethical approval was required.

\bibliographystyle{IEEEbib}
\bibliography{main_citations,refs}
\end{document}